\documentclass[runningheads]{llncs}
\usepackage{amsmath,amssymb,amsfonts}
\usepackage{graphicx}
\begin{document}
\title{Towards Smart Fake News Detection Through Explainable AI}
%
%
\author{Athira A B \inst{1} \and S D Madhu Kumar \inst{1}\and Anu Mary Chacko \inst{1}}
%
%
\institute{Department of Computer Science and Engineering \\  National Institute of Technology Calicut, Calicut \\
\email{athira.bala.sree@gmail.com \\ madhu@nitc.ac.in \\ anu.chacko@nitc.ac.in}}

\maketitle              
\begin{abstract}
People now see social media sites as their sole source of information due to their popularity. The Majority of people get their news through social media. At the same time, fake news has grown exponentially on social media platforms in recent years. Several artificial intelligence-based solutions for detecting fake news have shown promising results. On the other hand, these detection systems lack explanation capabilities, i.e., the ability to explain why they made a prediction. This paper highlights the current state of the art in explainable fake news detection. We discuss the pitfalls in the current explainable AI-based fake news detection models and present our ongoing research on multi-modal explainable fake news detection model.
\end{abstract}
\section{Introduction}
\begin{table*}[htbp]
\caption{Summary of existing explainable AI approaches for fake news detection}
\scalebox{0.9}{
\begin{tabular}{|p{0.1\linewidth}|p{0.2\linewidth}|p{0.2\linewidth}|p{0.2\linewidth}|p{0.4\linewidth}|}
\hline
\textbf{Research Articles } & \textbf{Features analysed} & \textbf{Methodology used} & \textbf{ Explainability approaches used } & \textbf{Accuracy} \\
\hline
\cite{b1} &  News attributes, Semantic meanings of news statements and Linguistic features of news statements & MIMIC, PERT, and ATTN frameworks & Mimic Learning (or Explanation by simplification), Perturbation based explanation, Self-attention based explanation and Visual explanation & 67.1\%, 67.3\%, 53.2\% accuracy for MIMIC, ATTN and PERT respectively. \\  \hline
\cite{b2}\cite{b3}  & News content and User comments &  RNN and Bidirectional GRU & 
Co-attention captures intrinsic explainability and Visual explanation & Outperforms related works by at least 5.33\% \\  \hline
\cite{b4} &  News content and User comments(filtered) & Multi nominal Knowledge base and RNN & Co attention Mechanism & Accuracy increased above 90\% for all machine learning algorithm\\  \hline
\cite{b5} & Interpretable textual features &  CoCoGen, BRNN, and GRU & Ablation explanations & Accuracy
99\% accuracy for ISOT and 27.4\% for LIAR dataset

 \\  \hline
\cite{b6} & Textual content & BiDir-LSTM-CNN and BERT &  Attention based and Visual explanation & Accuracy
Achieved 85\% of accuracy \\  \hline
\cite{b7} &  Network based patterns  & Supervised learning algorithms with five-fold cross-validation &  Empirical studies on social psychological theories & 93\% accuracy for politifact and 86\% for Buzzfeed data set\\  \hline
\cite{b8} &  News contents & Tstelin Machine (TM) &  Local and Global interpretability & Outperforms previous works by at least \%  \\  \hline
\cite{b9} &  Semantic contents of news articles &
 Knowledge Graph Framework (KGF) & Structured explainability (Relational Level)& 71.4\% , 86.0\%, 73.3\% accuracy for data sets Celebrity, PolitiFact and GossipCop respectively.\\  \hline 
\cite{b10} &  source tweet and retweet users & Graph Attention Network & Attention based explanation & Outperform state-of-the-art methods by 2.5\% in accuracy.\\  \hline
\cite{b11} & user-based, textual and temporal features in propagation pattern &  propagation network &  Attention based explanation & Outperformed 89.2\% for dataset Gossipcop\\  \hline
\cite{b16} & user-based and  textual  features & GCAN &  Attention based explanation & Outperformed state-of-the-art methods by 16\% in accuracy \\  \hline
\end{tabular}}
\end{table*}
Artificial intelligence (AI) is a technology which has the potential to change changes every area of the society. These systems analyze data and employ computer programs to imitate human thinking and actions. AI relieves humans from repetitive jobs and assists them with cognitive skills such as decision-making and problem-solving. Computers can capture large volumes of data using AI and utilize their learned intelligence to make optimal decisions and discoveries in seconds than humans.  \\

Social media platforms like Facebook, Instagram, and Twitter have become an integral part of our daily lives. These platforms use AI-based tools to monitor and manage social media content. Social media platforms have adopted AI-based tools as a backbone to protect both platforms and users alike. \\

As social media has grown in popularity, people's primary source of information has switched to internet news. Meanwhile it has also paved way for the propagation of fake through the social media. Fake news has become a threat to businesses and academics alike. Initially, to mitigate the fake news dissemination, manual methods were used. These methods include the use of fact-checking websites and crowd-sourcing human employs. Both of these approaches compare true news posts against unverified news. However, human fact-checking is futile because a large amount of data is generated every day and spreads quickly. The difficulties created by manual fact-checking cleared the door for automatic identification of fake news. \\

Fake news spread has various aspects that make it difficult to solve. The most challenging aspect in  controlling and detecting the spread of fake news is identification of the constituents of fake news. The term ``fake news" has no uniform definition.\\

``Fake news is defined as false or misleading information presented as news, Its goal is frequently to harm a person's or entity's reputation or to profit from ad revenue"\cite{b15}.\cite{b18} gave two definitions for fake news: “Fake news is false news,” is the  broad definition \cite{b18}, and “fake news is intentionally false news published by a news outlet”-  is the  narrow definition from \cite{b18}.\\

Based on the literature review we find that, fake news detection is modelled traditionally as a binary classification problem. We can define it as follows: \\

Let x be the features extracted from the social media news posts, a$_{i}$ represents the features extracted from the news item x. where i=1,2,3,$\dots$n. The fake news detection model \textit{F} classify the news item as real or fake. \\
\begin{equation}
  F(a_{i}) =
  \begin{cases}
            0 & \text{if x is a fake news} \\
            1 & \text{otherwise} \\
  \end{cases}
\end{equation}
Depending on the number of the features a$_{i}$ used by the fake news detection model \textit{F}, the fake news detection problem is categorised into single-modal or multi-modal. In single-modal fake news detection, only a single type of feature is analysed. In the case of multi-modal fake news detection, various feature combinations are examined. \\

Majority of the research works from the literature use AI-based methods to solve the fake news detection problem. Various machine learning and deep learning algorithms were used to achieve promising results. Deep learning based methods achieved more accurate results than the naive machine learning approaches. Even though black-box models are giving promising results, explainability (why a news piece is classified as fake news) is missing. Providing explanations for the prediction gives the transparency. Improving transparency help's users to avoid over trusting the fake news detection algorithms.This shows the significance of explainability in fake news detection systems. However, work in this area of explainable fake news detection is limited. \\

The rest of the paper is outlined as follows: Section 2 highlights the application of explainable AI (XAI) to fake news detection, as well as related research in this area. Our work on visual and textual content analysis of news stories on social media platforms is described in Section 3. Section 4 concludes the paper.

\section{Explainable AI (XAI) applied to Fake News Detection}
People are spending more time on social media platforms to receive the latest news and information, and their role in the spread of fake news cannot be neglected. The motivation behind fake news creation and spreading can range from financial gain to criminal or political benefits. In order to spread fake news on social media, high-quality tools for purchasing followers, likes, and supporting  comments on posted news items are available. The social media networks have a hard time detecting them.\\

Recently, many researchers have come up with AI-based solutions to fight against fake news dissemination. Several models are solved using large data sets and complex machine learning models. Machine learning models may pick up biases from the training data. Trusting the model is purely based on classification accuracy, ignoring the reason why a certain prediction was made, affects the performance of the model. It is easier to trust a system that explains its predictions compared to complex machine learning models. \\

Many researchers recently applied explainable AI for the detection of fake news. Table 1 summarises state of the art methods that applies explainable AI for fake news detection.\\

When going over the table summarized we found that most of the existing efforts in explainable fake news detection, explore news content features for providing explanations. Only \cite{b2} , \cite{b3} and  \cite{b4} derived explanation from the perspective of news content and social engagements. Different approach was by \cite{b7}, where they make use of patterns identified in friendship networks to build an explainable fake news detection. Recently, \cite{b11} proposed a methodology for explainable fake news identification that leverages user-based, textual, and temporal characteristics in the dissemination pattern. Even though various techniques have recently combined visual and textual features to capture distinct aspects of fake news, we couldn't find any study that use visual features in explainable fake news identification.\\
We are currently working on a visual feature based explainable fake news detection model. The details of the model is discussed in the following section.

\section{Visual and Textual Content Analysis}
The current research work in XAI applied to fake news detection mainly focuses on the textual feature analysis of the news content. Such approaches will succeed only if the news items follow a common writing style. Once the writing style is changed, this method fails in giving better accuracy. The addition of other features like social engagement, user profile information, and visual features helps to solve this problem.\\

Fake news posts containing multimedia information are more appealing to social media users than text-only posts. These visual elements can help to increase the accuracy of fake news detection methods. Our work concentrates on visual and textual content analysis of news articles on social media sites. The link between the textual and visual elements is employed to describe why the prediction was made. The work is progressing well and the initial results are encouraging. \\
  
`The high level design and the details of the work flow of our approach is shown in Fig 1. Both the textual and visual features extracted are fed into the deep learning model. Prediction of news into fake or real is done here. An additional module is employed for providing explanation for the predicted output. FakeNewsNet dataset is used for experimentation. \\
\begin{figure}[htbp]
\centering\includegraphics [scale=0.5]{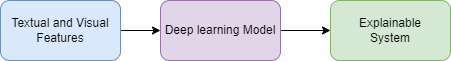}
\caption{work flow diagram}
\label{fig}
\end{figure} \\
\section{Conclusion}
We have emphasized the role of AI in the need for detecting and controlling the spread of fake news in social media in this paper. We have thrown light on how AI can help mitigate the challenges posed by the spread of fake news. The importance of explainable AI in the identification of fake news and our work on this is briefly reported here. We also presented an in-depth analysis of the recent ways for detecting fake news with the support of explainable AI. The initial results in our XAI-based fake news detection models are promising, and  we are  going ahead on the developments of techniques which can be used in multimedia fake news detection with satisfactory explanations.

\end{document}